\documentclass[letterpaper, 10 pt, conference]{ieeeconf}  

\IEEEoverridecommandlockouts                              

\usepackage{color}
\usepackage{caption}
\usepackage{graphics} 
\usepackage{amsmath} 
\usepackage{amssymb}  
\usepackage{biblatex} 
\usepackage{booktabs}
\usepackage{graphicx}
\usepackage[capitalise]{cleveref}
\title{\LARGE \bf
Enhancing Multi-Robot Perception\\via Learned Data Association
}

\author{Nathaniel Glaser, Yen-Cheng Liu, Junjiao Tian, and Zsolt Kira\\
Georgia Institute of Technology\\
\texttt{$\{$nglaser,ycliu,jtian73,zkira$\}$@gatech.edu}
}
\begin{document}
\maketitle


\begin{abstract}
In this paper, we address the multi-robot collaborative perception problem, specifically in the context of multi-view infilling for distributed semantic segmentation.  This setting entails several real-world challenges, especially those relating to unregistered multi-agent image data.  Solutions must effectively leverage multiple, non-static, and intermittently-overlapping RGB perspectives. To this end, we propose the Multi-Agent Infilling Network: an extensible neural architecture that can be deployed (in a distributed manner) to each agent in a robotic swarm.  Specifically, each robot is in charge of \textit{locally} encoding and decoding visual information, and an extensible neural mechanism allows for an uncertainty-aware and context-based exchange of intermediate features.  We demonstrate improved performance on a realistic multi-robot AirSim dataset.
\end{abstract}


\section{INTRODUCTION}

Multi-agent robotics is a burgeoning sub-field.  Thanks to continued advances in robotic hardware, individual robots within a robotic swarm are now afforded richer sensors, more powerful processors, and stronger connectivity than ever before.  Specifically, swarm hardware has evolved such that individual robots are now able to \textit{locally} perform some of the expensive computations associated with processing high-dimensional sensor data such as images, especially using efficient deep neural networks.  

This paper tackles the real-world setting where each robot performs local computations and inference (\textit{e.g.} feature extraction and semantic segmentation) but also has the ability to combine its local observations with information from other robots to improve performance.  Specifically, we focus on the task of distributed, multi-robot semantic segmentation.  Namely, a degraded robot must generate an accurate semantic mask despite missing information.  To achieve this end, each individual robot is allowed to communicate with its peers, but must perform computations \textit{locally}.  Unlike other problem settings, there are several challenges inherent in this task, including occlusions, degradations, and intermittent partially-overlapping (or non-overlapping) views.

\begin{figure}
\centering
\includegraphics[width=\linewidth]{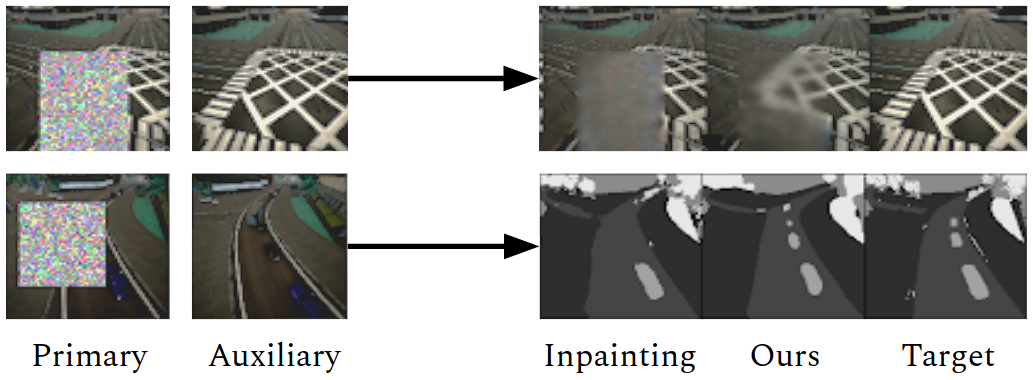}
\caption{\textbf{Qualitative results of our model on the collaborative perception task}.  The primary robot must overcome local image degradation by querying, corresponding, and exchanging features with the auxiliary robot(s).  Our model enables information sharing for any distributed perception task that uses an intermediate spatial feature map, as with reconstruction (top) and semantic segmentation (bottom).}
\label{fig:Qualitative_Infill}%
\end{figure}

In this paper, we address these challenges by combining learning-based methods across several fields, including visual odometry~\cite{fischer2015flownet,ilg2017flownet,Sun_2018_CVPR,xu2017accurate,wang2018occlusion}, SLAM~\cite{avraham2019empnet}, and multi-view fusion~\cite{su2015multi}.  While each of these methods have been developed for simpler, more specialized settings, our combined approach advances towards the general swarm perception challenge.  Unlike visual odometry and SLAM, we deal with the problem of non-sequential and large-displacement inputs, distributed information sharing, as well as local vision-based inference tasks.  Unlike common multi-view methods, we directly perform learned data association and alignment to deal with irrelevant views and degradations.  Furthermore, unlike inpainting methods, we generate outputs that accurately represent the scene, rather than producing ``reasonable'' hallucinations, as shown in Fig.~\ref{fig:Qualitative_Infill}.  

We summarize the contributions of our paper as follows:
\begin{itemize}
    \item We introduce a variant on the \textit{Multi-Robot Collaborative Perception} task~\cite{liu2020when2com, liu2020who2com} in which an individual robot must collaborate with supporting robots to overcome partial image occlusion and degradation.
    \item We propose an end-to-end learn-able \textbf{M}ulti-\textbf{A}gent \textbf{I}nfilling \textbf{N}etwork, \textbf{MAIN},  that (1) extracts spatial features for pairwise comparison, (2) leverages spatial \textbf{context} and matching \textbf{uncertainty} to produce a smoothed correspondence volume, and (3) uses this volume to sample and fuse features from supporting robots.  Each of these components helps the network infill missing information and ensures that the final per-agent perception output (such as a semantic mask or reconstructed input) is less vulnerable to partial input degradation.
    \item We test our network on  \textit{distributed} semantic segmentation for multiple RGB observers, where each robot can improve its local perception corresponding and sharing information.  We demonstrate superior performance of our model compared against several baselines in a photo-realistic multi-robot AirSim environment.
\end{itemize}

\section{RELATED WORK}

Our work draws inspiration from several domains, including image correspondence, optical flow, visual odometry, simultaneous localization and mapping, and multi-view learning.  We specifically highlight the deep learning approaches within each domain.

\textbf{Image Correspondence} focuses on identifying which features (or pixels) in one frame represent the same reprojected feature in another frame.  For our swarm setting, this general procedure is essential for identifying which features to exchange between robots.  Modern work in this domain uses deep convolutional networks to generate features that are suitable for matching with procedures that use learned interest points~\cite{detone2017toward,detone2018superpoint}, patches~\cite{rocco2017convolutional}, and cost volumes~\cite{melekhov2019dgc}.

\textbf{Optical Flow} and \textbf{Visual Odometry} enforce additional constraints on the image correspondence setting.  Most notably, they often assume that images are captured sequentially and with small relative displacement.  These assumptions are typically baked into the architectures of optical flow models~\cite{fischer2015flownet,xu2017accurate,Sun_2018_CVPR} and visual odometry models~\cite{detone2016deep,wang2017deepvo}.  More recent work~\cite{ilg2017flownet} has addressed larger displacement settings, though at the expense of added architectural and computational complexity.

\textbf{SLAM} incorporates streams of odometry and sensor information to generate a map and localize within it.  This map can be extended with additional measurements, trajectories, and robots.  Most notably for this work, EMPNet~\cite{avraham2019empnet} corresponds learned image features between current and past observations and then subsequently performs alignment between these embeddings to maintain a geometrically consistent ``map''.  

\textbf{Multi-View Learning} uses numerous perspectives to build a shared representation that is subsequently decoded towards some task.  For instance, the MVCNN architecture~\cite{su2015multi} performs view-pooling on the high-dimensional learned features from several views to perform 3D object classification.

Similar to prior work, our work uses a cost volume as the mechanism for corresponding learned features.  Additionally, we extend the cost volume decoder of DCG~\cite{melekhov2019dgc} to handle partial occlusions and image degradations.  By explicitly training the decoder to use spatial \textit{context} and matching \textit{uncertainty} to infill a warping mask (and not simply infill the raw input image), we demonstrate resilience to significant viewpoint shifts and degradations.  We further show that the infilled warping mask can be used to sample features from a supporting agent.  Finally, we extend our work to satisfy the \textit{distributed} multi-robot setting, where each robot performs local computations and engages in optional and opportunistic information exchange between other robots.
\section{METHOD}

\begin{figure*}
\centering
\includegraphics[width=\linewidth]{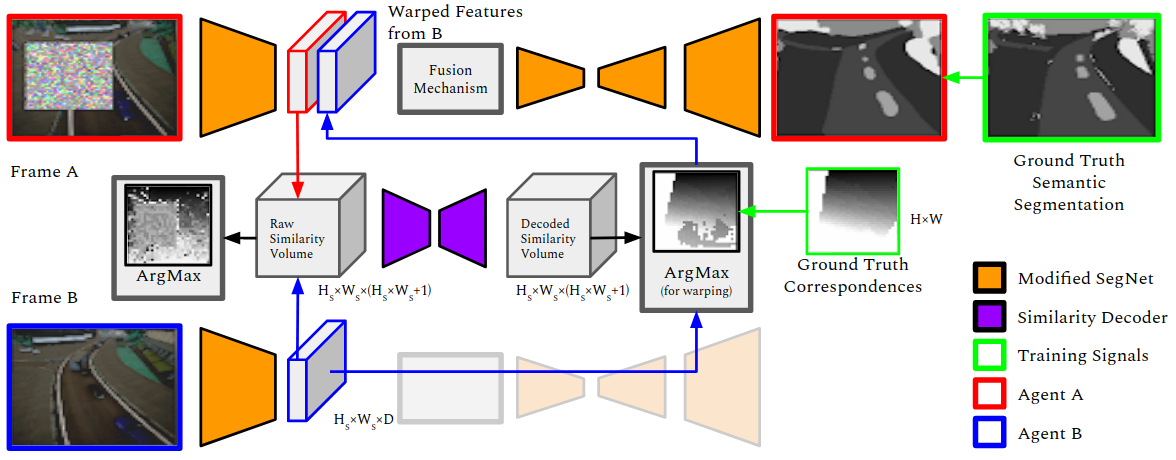}
\setlength{\belowcaptionskip}{-15pt}
\caption{\textbf{Overall architecture for Multi-Agent Infilling Network}.  We pass each robot's images through the first few layers of a SegNet architecture.  We then perform a dense pairwise comparison between the resulting spatial feature maps (from each agent) to generate a raw similarity volume, and we use an auto-encoder network to smooth and infill it.  With this decoded similarity volume, we sample (or warp) features from the auxiliary network to supplement the task decoding of the primary network.  Note that our method extends to any number of distributed agents.}
\label{fig:overall}
\end{figure*} 

\subsection{Multi-Agent Infilling Network}\label{sec:MAIN}
The multi-robot collaborative perception task entails several real-world challenges, many of which stem from handling unregistered swarm image data.  Namely, we must deal with multiple, non-static, and intermittently-overlapping RGB perspectives.  For the specific task of distributed semantic segmentation, we must enhance the segmentation output of a primary robot (with a degraded input) by collaborating with several other robots (with non-degraded inputs).  In order to properly handle this scenario, we must design an architecture that can efficiently summarize, correspond, and exchange information in a distributed manner.

As such, we propose a neural architecture that extends a generic convolutional encoder-decoder structure.  Specifically, we insert a learned network module a few layers deep into a backbone convolutional network (at the spatial feature level).  This inserted module uses the intermediate spatial feature maps from each robot to perform data association, smoothing, and feature exchange.  After features have been associated and exchanged, the remaining convolutional layers decode this information into the target output (\textit{i.e.} reconstruction or segmentation).  Our procedure is summarized in Figure~\ref{fig:overall} and further detailed in the following sections.

\textbf{SegNet Backbone} We use the SegNet architecture~\cite{badrinarayanan2017segnet} as the backbone for our multi-robot infilling network.  We simply insert our correspondence module into a single layer of this backbone.  Specifically, the SegNet layers that precede our module serve as a spatial feature encoder, and the layers that follow it serve as the task-specific decoder.  This design allows for each robot to compute its own encoding and decoding in a distributed manner.  We use a  VGG16~\cite{simonyan2014very} encoder pretrained on ImageNet~\cite{krizhevsky2012imagenet}. 

\textbf{Data Association} The SegNet encoder ingests a raw RGB image of size $H \times W \times C$ and produces a spatial feature map $f$ of downsampled size $H_s \times W_s \times K$.  Using the intermediate spatial features from two robots (denoted with the subscripts $A$ and $B$), we compute a dense pairwise distance volume:
\begin{equation}
  \mathcal{D}_{AB}[x,y,x',y'] = d(f_A[x,y],f_B[x',y']),
\end{equation}
where $d$ defines a distance metric ($L_2$) between two feature vectors of dimension $K$.  After computing each entry, we then rearrange this 4D similarity volume into a 3D tensor of size $H_A \times W_A \times(H_B \times W_B)$, which describes how each spatial feature from Robot $A$ matches to all spatial features from Robot $B$.  Additionally, we compute a ``no-match'' score:
\begin{equation}
  \mathcal{D}_{A\varnothing}[x,y] = d(f_A[x,y],\vec{0}).
\end{equation}
The added score gives the network a mechanism for identifying when the features in question are not distinctive enough for pairwise matching, as used in prior work~\cite{avraham2019empnet}.  We append this ``no-match'' tensor ($H_A \times W_A$) to the back of the 3D similarity tensor, which gives a final volume of $H_A \times W_A \times(H_B \times W_B + 1)$.  Finally, we compute the softmax distribution across the channel dimension of the \textit{negated} 3D volume, yielding a normalized distribution of matching scores between patches and of the no-match case.  In essence, the SegNet encoder provides a high-dimensional representation of spatial regions (patches) within the original image, and the similarity volume computes the visual similarity between each of these regions.

\textbf{Uncertainty-Aware and Context-Based Smoothing} Since we compute pairwise matching scores between \textit{single} spatial feature cells, the resulting similarity volume is quite noisy.  We therefore pass the raw similarity volume through a convolutional encoder-decoder architecture, which has access to both matching uncertainty (from the probability distribution in the channel dimension) and spatial context (from the preserved spatial arrangement).  It outputs a smoothed correspondence volume that can be used for more reliable indexing and weighting of the contributing feature maps.

\textbf{Cross Feature Sampling and Fusion} Given a smoothed correspondence volume of size $H_A \times W_A \times(L_B \times W_B + 1)$, we take the ArgMax of this volume across the channel dimension to generate a 2D mapping of each spatial cell in robot $A$ to the best matched cell in robot $B$.  We use this correspondence map to sample the spatial features from robot $B$.  After sampling, we now have two (aligned) spatial feature grids--the original spatial features from $A$ and the aligned features from $B$.  Based on which per-feature similarity score is higher (\textit{i.e.} the ``no-match'' score or  the best match score between $A$ and $B$), we use hard selection to forward that feature to the remainder of the SegNet network, which decodes the spatial features into the target output format.

\textbf{Multi-Robot Extension} For brevity, we described our data association and feature sharing procedure for only \textit{two} robots.  However, our procedure generalizes to the \textit{multi}-robot case.  In this case, each robot performs local encoding and decoding, but each also transmits its intermediate features to any number of neighbors.  In turn, each neighbor computes the matching scores and alignment of all incoming feature maps, and each subsequently chooses the best feature for final decoding.  The general framework is extensible--each robot can handle any number of incoming feature maps, potentially improving performance.
\begin{table*}
\resizebox{\linewidth}{!}{%
\begin{tabular}{lc|c|cccc||c|c|cccc}
\toprule
&  \multicolumn{6}{c}{\textit{\textbf{Sequence}} Split} & \multicolumn{6}{c}{\textit{\textbf{Cross}} Split} \\
\midrule
& Mean & Mean & \multicolumn{4}{c}{Class IoU} & Mean & Mean & \multicolumn{4}{c}{Class IoU} \\
& Accuracy & IoU & Building & Bus & Car & Road & Accuracy & IoU & Building & Bus & Car & Road \\
\midrule
Inpainting   &  37.08   &   31.17   &   74.34   &   0.00   &   5.52    &   83.90 &   32.69   &   25.91   &   66.46   &   5.15   &   5.77   &   77.95 \\
InputStack    &  35.75   &   29.20   &   61.40   &   1.34   &   2.20   &   77.44 &  41.44   &   33.68   &   66.59   &   5.80    &   7.54    &   \textbf{84.02} \\
FeatureStack    &   37.55   &   32.95   &   77.59   &   2.64    &   6.17    &   79.61 &   32.85   &   23.62   &   47.03   &   8.80   &   2.566   &   64.04 \\
ViewPooling &   36.09   &   30.48   &   74.58   &   0.00    &   9.32    &   73.44 &   22.57   &   13.35   &   33.88   &   0.00    &   0.00    &   42.41 \\
\midrule
NoSimLoss   &   30.25   &   22.69   &   59.38   &   0.00   &   0.00   &   68.59 & - & - & - & - & - & - \\
NoSmoothing  &   34.60    &   28.70  &   65.25   &   0.00   &   4.49   &   78.91 & - & - & - & - & - & -  \\
OneHot  &   48.40   &   45.04   &   81.80   &   23.10   &   \textbf{33.51}   &   87.75 & - & - & - & - & - & -  \\
\midrule
MAIN &   \textbf{52.83}   &   \textbf{46.34}   &   \textbf{83.10}   &   \textbf{30.55}   &   32.08   &   \textbf{88.86} &   \textbf{51.34}   &   \textbf{42.77}   &   \textbf{75.27}   &   \textbf{22.76}   &   \textbf{17.07}   &   83.88 \\
\bottomrule
\end{tabular}}
\setlength{\belowcaptionskip}{-15pt}
\caption{\textbf{Segmentation Results within Occluded Region on Single-Robot \textit{Sequence} and Multi-Robot \textit{Cross} Datasets}.}
\label{tab:combined_partial}
\end{table*}

\section{EXPERIMENTS}
\textbf{Dataset}. 
For our dataset, we use the photo-realistic, multi-robot AirSim~\cite{airsim2017fsr} simulator to collect synchronized RGB, depth, semantic segmentation, and pose information from a realistic city environment with dynamic objects.  Within this environment, we control a non-rigid swarm of 6 drones to various road intersections.  En route to these intersections, the drones rotate at different rates, ensuring that their views are dynamically and non-trivially overlapping.  Once the swarm arrives at an intersection, each drone randomly explores the region, further ensuring interesting frame overlaps.  With the collected information (and the intrinsic camera parameters), we then generate the ground truth pixel correspondences for (A) consecutive frames of each robot and (B) all image pairs between any two robots.  This dense pairwise data is useful during training and for evaluation, but it is unnecessary during actual deployment.  All images are $128 \times 128$ pixels.

We create two splits from this general AirSim dataset.  The \textbf{Sequence} split includes temporally-consecutive pairs of images collected \textit{in sequence} from each robot, whereas the \textbf{Cross} split includes temporally-aligned pairs of images collected \textit{across} pairs of robots.  The \textbf{Sequence} split is a simpler test case with less drastic viewpoint shifts, akin to a visual odometry dataset.  The \textbf{Cross} split is the more difficult \textit{multi-robot} case ($N=6$) where viewpoints are drastically different (and oftentimes, non-overlapping).  For both splits, we simulate occlusion by overlaying a rectangular area of uniformly-distributed noise on the primary image observation.  This specific degradation highlights the challenging case where important scene information is obstructed by a foreground obstruction, such as rain or dirt on a camera lens.

\textbf{Training.} For our experiments, we train the \textbf{MAIN} network with two ground truth signals: (1) the perception task signal and (2) the pixel correspondences between all pairs of robots.  We use a cross entropy loss for both the semantic segmentation and correspondence signals.  Additionally, though the network is trained in a dense, centralized manner, it can be deployed in a distributed manner.  We emphasize that, during inference, the only inputs to the network are a single RGB image per robot.  Additionally, we train and test several ablated \textbf{MAIN} variants and SegNet-based baselines. 

\textbf{Baselines.} We use several methods as baselines:\label{sec:baselines}

\textit{Inpainting} performs semantic inpainting based on the primary (degraded) frame only.  A standard SegNet architecture is trained to generate a non-degraded semantic segmentation mask from a single degraded input image, as inspired by~\cite{cai2017blind}.

\textit{InputStack} concatenates the raw RGB images from the primary and auxiliary views together as input, prior to a standard SegNet architecture.  This procedure resembles that of Deep Homography and Visual Odometry methods~\cite{detone2016deep,wang2017deepvo}.

\textit{FeatureStack} concatenates the feature encodings from the primary and auxiliary frames, after the first two SegNet convolutional downsampling blocks.

\textit{ViewPooling} performs view pooling on the (unwarped) feature encodings from the primary and auxiliary frames, similar to that of MVCNN~\cite{su2015multi}. 

\textbf{Ablated Variants.} We ablate our method in several ways:\label{sec:ablated}

\textit{NoSimLoss} is identical to the \textbf{MAIN} architecture described above, but it does not have a direct training loss for the similarity volume.  We include this model to show the necessity for ground truth correspondences, which may be obtained through accurate pose and depth information.

\textit{NoSmoothing} bypasses the smoothing network and directly uses the ArgMax of the raw matching scores to sample the auxiliary feature map.  We include this model to highlight the necessity of the context-aware smoothing network.

\textit{OneHot} removes valuable information from the raw similarity volume by converting it into a one-hot volume of best matches (prior to the smoothing network).  Since we hypothesize that the smoothing network uses the full per-pixel distributions to perform uncertainty-aware smoothing, we include this ablated variant to show what happens when that distribution information is suppressed.  

\textbf{Results.}\label{results}
We summarize our findings on the \textbf{Sequences} and \textbf{Cross} datasets in Table~\ref{tab:combined_partial}.  In both datasets, we observe that the \textbf{MAIN} architecture significantly outperforms all other models, especially with respect to segmentation accuracy on several noteworthy dynamic object classes (\textit{e.g.} bus, car, and truck).  In contrast, the other network variants are unable to capture these dynamic objects and have a reduced (but not entirely suppressed) ability to classify static objects.

Specifically, the \textbf{Inpainting} model produces surprisingly accurate segmentation masks, despite not having access to the auxiliary view.  Its non-zero score likely results from the strong static scene prior provided by the dataset environment; a standard SegNet network can use this strong prior to infill static portions of the scene.  On the other hand, the \textbf{InputStack}, \textbf{FeatureStack}, and \textbf{ViewPooling} models have access to the auxiliary view, but in fact, they suffer from a decrease in performance despite this extra information.  Since these models do not know the (ever-changing) alignment between the primary and auxiliary views, the auxiliary information corrupts the segmentation results.

Regarding the ablated variants of the \textbf{MAIN} architecture, we observe that the \textbf{NoSimLoss} and \textbf{NoSmoothing} models yield poor segmentation results.  These poor results motivate several of our architectural decision, such as including a direct correspondence loss and an intermediate smoothing network.  Additionally, the \textbf{OneHot} model shows that consolidating the per-cell matching distributions into a one-hot tensor of best matches causes a slight drop in performance.
\section{CONCLUSION AND FUTURE WORK}
We propose a variant on the multi-robot collaborative perception task where robots must work together to overcome a localized image degradation.  We address this task via a neural network architecture that uses \textit{context-based} and \textit{uncertainty-aware} data association to exchange features within a standard segmentation network.  We show that our model is able to (1) learn embeddings that are suitable for cross-robot matching, (2) generate a dense correspondence volume from these embeddings and perform context-based smoothing, and (3) use the smoothed correspondence to exchange features that are decoded to satisfy the final task.

Future work in this domain will involve improving the performance and efficiency of dense feature exchange, in a similar vein as ~\cite{liu2020who2com}.  Namely, bandwidth is expensive for a fully-connected multi-robot scenario, and exchanging full images or high-resolution spatial feature maps across \textit{all} robots may be prohibitive.

\section{Acknowledgement}
\label{sec:acknowledgement}
\noindent This work was supported by ONR grant N00014-18-1-2829.

\clearpage
\printbibliography
\end{document}